\documentclass{llncs}

\usepackage[T1]{fontenc}

\usepackage{graphicx}
\usepackage{amsmath,amssymb}

\usepackage{booktabs}
\usepackage{multirow}
\usepackage{tabularx}
\usepackage{array}
\usepackage{threeparttable}
\usepackage{colortbl}

\usepackage{xcolor}
\usepackage{microtype}
\usepackage{float}
\usepackage{hyperref}

\hypersetup{hidelinks}

\renewcommand{\arraystretch}{1.12}

\begin{document}

\title{\textsc{AlignUS}: MRI-Guided Ultrasound Representation Learning for ALS Classification from Tongue Images}
\titlerunning{\textsc{AlignUS} for ALS Classification}

\author{
{\normalsize
Khadijetou Abdel Ghader\textsuperscript{1} \and
Emani Babe\textsuperscript{1} \and
Lorenzo Pettinari\textsuperscript{2} \and
Meya Haroune\textsuperscript{1} \and
Sidaty El Hadramy\textsuperscript{2}
}
}

\authorrunning{K. Abdel Ghader et al.}

\institute{
{\small
\textsuperscript{1}Unité de Recherche en Systèmes Intelligents Avancés (URSIA), Institut Supérieur du Numérique (SupNum), Nouakchott, Mauritanie\\
\email{\{21016, 21068, meya.haroune\}@supnum.mr}
\\[2pt]
\textsuperscript{2}University of Basel, Basel, Switzerland\\
\email{\{lorenzo.pettinari, sidaty.elhadramy\}@unibas.ch}
}
}

\maketitle

\begin{abstract}
Amyotrophic lateral sclerosis (ALS) is a progressive neurodegenerative disease in which early assessment remains challenging, particularly in low-resource settings where MRI is often unavailable. High-resolution ultrasound (HRUS) of the tongue offers a portable and low-cost alternative for evaluating bulbar involvement, but learning reliable diagnostic models is limited by small datasets and the difficulty of extracting robust representations from ultrasound alone. We propose \textsc{AlignUS}, a cross-modal knowledge distillation framework that transfers anatomical knowledge from MRI to a HRUS-based classifier while requiring only HRUS at inference time. The model combines classification loss, supervised contrastive learning, and feature-level distillation to align HRUS representations with MRI embeddings. \textsc{AlignUS} achieves a patient-level balanced accuracy of 0.958, macro-F1 of 0.963, and ROC-AUC of 0.990, aggregated across four patient-level cross-validation folds, with consistent improvements over HRUS baselines and cross-modal alternatives. These results demonstrate that MRI-derived supervision can substantially improve ultrasound-based ALS assessment while preserving low-cost, inference-time independence from MRI.
\end{abstract}


\section{Introduction}

Amyotrophic lateral sclerosis (ALS) is a progressive neurodegenerative disease characterized by the degeneration of both upper and lower motor neurons. As the disease progresses, patients gradually lose voluntary motor control, typically evolving from focal muscle weakness to widespread functional impairment, bulbar dysfunction, respiratory decline, and eventually respiratory failure \cite{brown2017,kiernan2011,hardiman2017}. Median survival after diagnosis is approximately two to five years \cite{kiernan2011,sabatelli2014}, highlighting the importance of early and reliable diagnosis to enable timely clinical management, appropriate supportive care, and potential therapeutic interventions. Currently, ALS diagnosis remains primarily clinical and relies on a combination of neurological examination, electrophysiological assessment, and neuroimaging to exclude alternative conditions. Magnetic resonance imaging (MRI) is frequently used as part of the diagnostic workflow because it provides detailed structural information and enables the assessment of central nervous system abnormalities associated with motor neuron degeneration. However, ALS presents considerable diagnostic challenges due to its clinical heterogeneity, the overlap of early symptoms with other neurological disorders, and the absence of a single definitive biomarker \cite{arthur2016,turner2009}. Consequently, many patients experience prolonged diagnostic delays, often exceeding one year before reaching specialized care centers, with these delays particularly pronounced in low-resource healthcare settings where access to advanced imaging technologies such as MRI is limited by high acquisition costs and the need for specialized personnel \cite{Falcao_de_Campos2022}.

The broader literature on AI-assisted neurodegenerative disease assessment offers useful direction for addressing these limitations. Deep learning methods have achieved strong performance in MRI-based classification of motor neuron disease by learning complex patterns of neurodegeneration directly from imaging data \cite{litjens2017survey,shen2017deep,cheplygina2019notsosupervised}, and related work has explored fusing structural MRI with diffusion and functional imaging to capture complementary disease signatures \cite{ngiam2011}. Methods such as SF2Former \cite{kushol2023}, when trained on MRI, illustrate the capacity of these models to extract diagnostically relevant features from volumetric neuroimaging. Yet, these MRI-based AI methods still face MRI's cost and infrastructure barriers, motivating the search for more accessible imaging alternatives.

High-resolution ultrasound (HRUS) of the tongue represents a promising candidate for this purpose. The tongue is particularly relevant because bulbar dysfunction is common in ALS and directly affects essential functions, including speech, swallowing, nutrition, and airway protection. Hensiek et al.~\cite{hensiek2020} demonstrated that both 3T MRI and HRUS can capture structural changes in the tongue associated with bulbar impairment and disease progression, and related work has further shown that tongue ultrasound measurements can reveal structural alterations associated with bulbar involvement in ALS \cite{schreiber2020tongue}. Compared with MRI, HRUS is portable, inexpensive, non-invasive, and suitable for repeated examinations, making it particularly attractive for monitoring disease progression and extending ALS assessment to settings where MRI access is limited. Despite these advantages, the use of tongue ultrasound for ALS assessment remains largely dependent on expert interpretation, limiting its scalability and reproducibility. AI-based methods could provide a means to support clinicians by enabling objective and consistent detection of disease-related ultrasound patterns. However, research on AI-based classification of ALS using tongue ultrasound remains limited, with existing studies constrained by the scarcity of annotated imaging datasets and the difficulty of collecting sufficient patient data for robust model development.

Beyond dataset scarcity, developing reliable AI models from HRUS is further complicated by operator-dependent acquisition variability and class imbalance between patients and controls, limiting models' ability to learn robust, generalizable representations and widening the gap between HRUS's clinical potential and its automated use.
To address these challenges, we propose \textsc{AlignUS}, a cross-modal knowledge distillation framework \cite{hinton2015distilling,gou2021survey} that uses MRI-derived representations during training to improve ALS classification from tongue ultrasound, requiring only HRUS at inference. Building on prior cross-modal distillation work in medical imaging \cite{gupta2016crossmodal}, \textsc{AlignUS} transfers complementary structural knowledge from MRI without depending on it at deployment, preserving the portability and low cost of ultrasound. The main contributions of this work are:
\begin{itemize}

\item To the best of our knowledge, we introduce the first framework to predict ALS from tongue HRUS images using cross-modal knowledge distillation from MRI, without requiring MRI during inference.

\item We design a framework adapted to low- and middle-income settings, relying solely on accessible, low-cost ultrasound imaging at deployment while benefiting from MRI-derived knowledge during training.

\item We conduct experiments on the dataset proposed by \cite{hensiek2020}, demonstrating state-of-the-art results for ALS assessment from tongue ultrasound images.

\end{itemize}

\section{Method}
\label{sec:method}

\begin{figure}[h]
    \centering
    \includegraphics[width=1\linewidth]{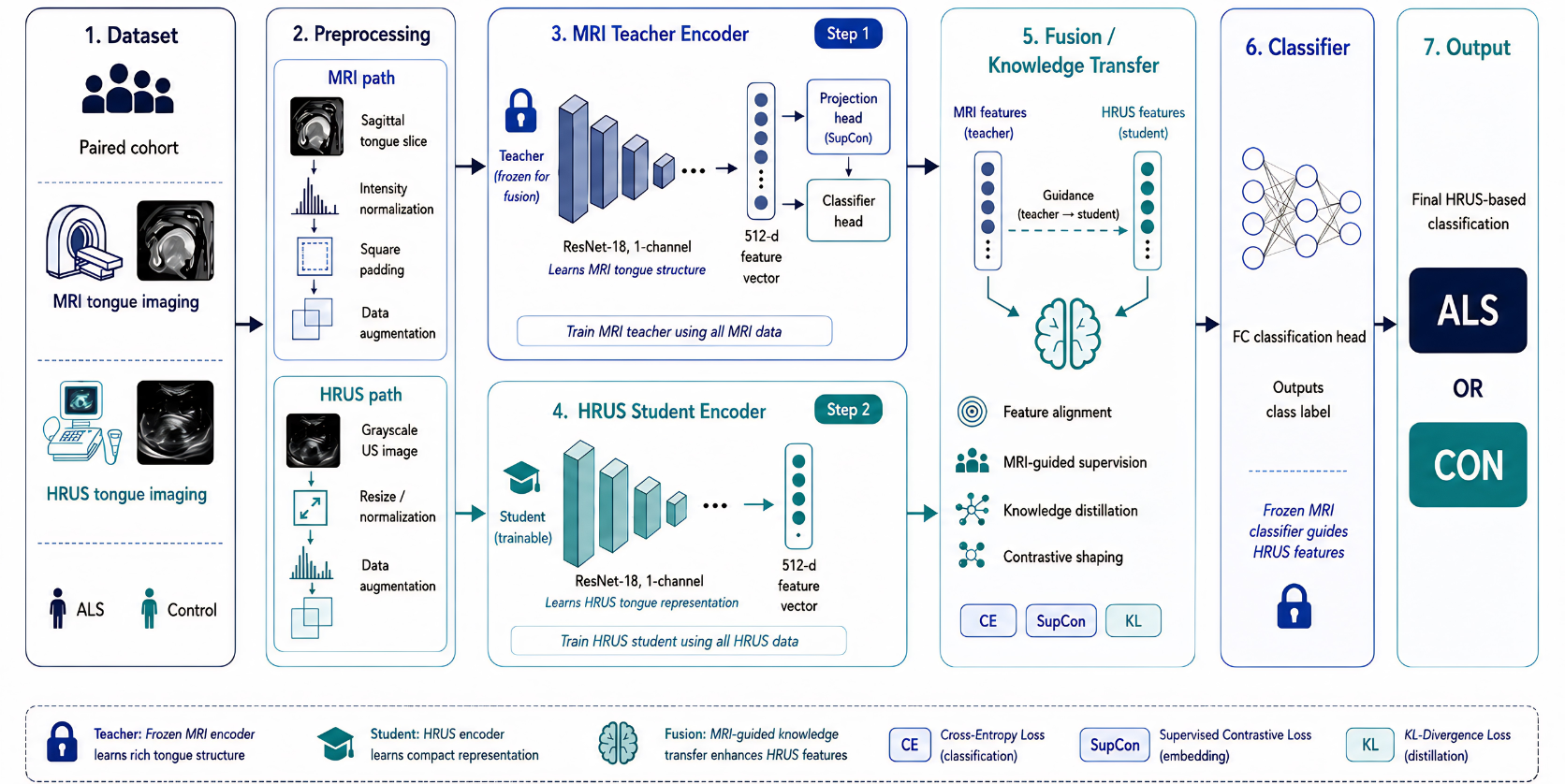}
    \caption{Overview of the proposed \textsc{AlignUS} framework. It follows a teacher--student design where an MRI encoder is first trained to learn representations from MRI data. The learned MRI encoder is then frozen and used to guide the training of an HRUS encoder through supervised classification, supervised contrastive learning, and knowledge distillation. During inference, only the HRUS encoder and classifier are used.}
    \label{fig:pipeline}
\end{figure}

We propose \textsc{AlignUS}, a teacher--student framework that leverages anatomical representations learned from MRI to improve HRUS-based diagnosis without requiring MRI at inference time. The central idea behind \textsc{AlignUS} is to use MRI as a modality during training: an MRI encoder first learns a robust representation of tongue anatomy and disease-related characteristics, which is subsequently transferred to an HRUS encoder through feature and prediction alignment. After training, only the HRUS encoder is retained, allowing the model to operate using ultrasound images alone. An overview of the proposed \textsc{AlignUS} framework is presented in Figure~\ref{fig:pipeline}.

\begin{figure}[h]
    \centering
    \includegraphics[width=0.95\linewidth]{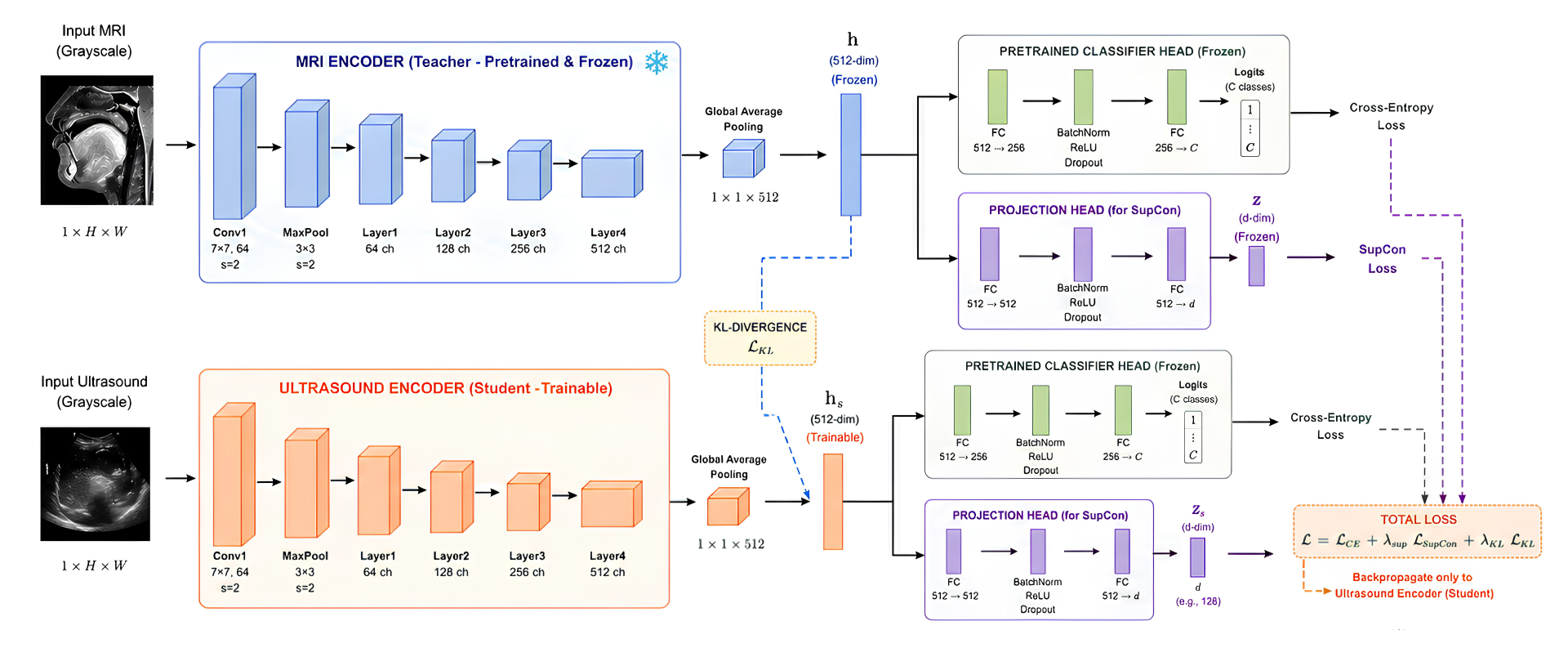}
    \caption{Detailed architecture of the MRI and HRUS encoders, along with the loss functions used to train the \textsc{AlignUS} framework.}
    \label{fig:architecture}
\end{figure}

As shown in Figure~\ref{fig:pipeline}, the framework consists of two stages. First, an MRI encoder is trained on labeled MRI data to learn a discriminative anatomical representation using classification and supervised contrastive objectives. The trained MRI encoder is then frozen and used as a teacher to guide the training of an HRUS student encoder through knowledge distillation. The student learns to approximate the MRI-informed feature space while relying only on HRUS images, enabling MRI-free inference. Figure~\ref{fig:architecture} provides a detailed overview of each stage of \textsc{AlignUS}, and the following sections describe its components.


\subsection{MRI Encoder}

The first component of \textsc{AlignUS} is the MRI encoder, which provides the anatomical reference representation used to guide HRUS learning. We chose ResNet-18 as the backbone architecture for both MRI and HRUS encoders. This architecture provides a good balance between representation capacity and robustness for the available dataset size, while residual connections improve optimization stability during contrastive representation learning. For MRI input, the first convolutional layer was adapted to accept single-channel grayscale images, while the remaining residual blocks were unchanged. For each MRI volume, we selected the single 2D sagittal slice with the highest image variance, under the assumption that greater spatial variation corresponds to better visualization of the tongue structure. The selected MRI slice was intensity-normalized using percentile clipping followed by z-score standardization and was square-padded to preserve the original aspect ratio during training. The MRI encoder produces a 512-dimensional embedding via global average pooling.  This embedding is then given as input to a linear classification head with dropout~\cite{srivastava2014dropout} for ALS/Control prediction and an MLP projection head that feeds a supervised contrastive objective. The contrastive formulation, following Khosla et al.~\cite{khosla2020supcon} and the SimCLR framework~\cite{chen2020simclr}, pulls embeddings from the same diagnostic class together while pushing those from opposing classes apart. Therefore, the MRI encoder is optimized using a combination of weighted cross-entropy and supervised contrastive loss: $\mathcal{L}_{\mathrm{MRI}} = \lambda_{\mathrm{CE}}\mathcal{L}_{\mathrm{CE}} + \lambda_{\mathrm{SupCon}}\mathcal{L}_{\mathrm{SupCon}}$.

\subsection{HRUS Encoder}

The second component of \textsc{AlignUS} is the HRUS student encoder, which learns to generate representations aligned with the MRI teacher while using ultrasound images as input. During this stage, the MRI encoder and classification head remain frozen, ensuring that the anatomical representation learned from MRI defines a stable target feature space. The HRUS encoder follows the same architecture as the MRI branch, using a single-channel ResNet-18 backbone with a 512-dimensional output after global average pooling. The architectural symmetry between the two modalities enables direct comparison between MRI and HRUS embeddings and allows the student representation to be aligned with the teacher without additional projection networks. To transfer the anatomical knowledge learned from MRI to HRUS, the student encoder is optimized using a multi-objective training strategy that combines supervised classification, supervised contrastive learning, and knowledge distillation. The overall HRUS training objective is defined as: $\mathcal{L}_{\mathrm{HRUS}} =
\lambda_{\mathrm{CE}}\mathcal{L}_{\mathrm{CE}}
+
\lambda_{\mathrm{SupCon}}\mathcal{L}_{\mathrm{SupCon}}
+
\lambda_{\mathrm{KL}}\mathcal{L}_{\mathrm{KL}}$.

The cross-entropy term provides direct supervision for ALS/Control classification. However, relying solely on classification loss may result in poorly structured representations, particularly when training with limited and imbalanced datasets. Therefore, supervised contrastive learning is incorporated to regularize the embedding space by reducing intra-class variation and increasing separation between diagnostic categories. The knowledge distillation term enables the transfer of MRI-derived information to the HRUS encoder. Specifically, the KL-divergence loss~\cite{hinton2015distilling,romero2015fitnets,park2019relational} encourages the student's predictions to match the soft output distribution of the frozen MRI teacher. This allows the student to benefit from the teacher's learned confidence patterns and inter-class relationships, which are not captured by hard diagnostic labels alone. Through this strategy, \textsc{AlignUS} follows the principles of cross-modal distillation~\cite{gupta2016crossmodal} and multimodal representation learning~\cite{ngiam2011}, where a modality available only during training guides the learning of a representation suitable for deployment in a single-modality setting.

\section{Results}
\label{sec:results}

\subsection{Dataset and implementation details}


The dataset consisted of publicly available tongue imaging data from Schreiber et al.~\cite{schreiber2020tongue}. After excluding incomplete cases, the final cohort comprised 62 subjects (48 ALS, 14 Controls) and 513 HRUS tongue images. MRI data were used exclusively for training the teacher model, while HRUS images were used for the student model. All images underwent standard preprocessing, including resizing and intensity normalization, and were split at the patient level prior to frame-level sampling into training, validation, and test sets to avoid data leakage. To address class imbalance, the training set was balanced via oversampling of control subjects using augmented samples, with geometric and photometric augmentations applied only during training. All experiments were implemented in PyTorch and trained on an NVIDIA A100 GPU. The MRI encoder was optimized using AdamW with a learning rate of $3\times10^{-4}$, weight decay of $1\times10^{-4}$, batch size of 8, cosine annealing for up to 50 epochs, and early stopping based on validation balanced accuracy. The HRUS encoder used the same optimizer settings, with loss weights set to $\lambda_{\mathrm{CE}}=1.0$, $\lambda_{\mathrm{SupCon}}=0.2$, and $\lambda_{\mathrm{KL}}=0.10$, selected via validation grid search. Model selection and evaluation were performed using \textbf{four-fold patient-level cross-validation}, ensuring that all images from a given subject were assigned exclusively to either the training, validation, or test fold and never shared across splits. In each fold, approximately 75\% of subjects were used for training, with the remaining subjects evenly partitioned into validation and held-out test sets; all the results reported in this work correspond to the aggregated performance across all four folds on the respective held-out test partitions.

\subsection{Quantitative results}


Table~\ref{tab:classification_report} shows that, at the class level, \textsc{AlignUS} achieves perfect recall for ALS (1.00), indicating that all ALS patients in the test cohort are correctly identified, alongside strong precision (0.94) and F1-score (0.97). Performance on the control class is also strong, with precision of 1.00, recall of 0.92, and F1-score of 0.95. 


\begin{table}[h]
\centering
\caption{Patient-level classification report on the held-out test set.}
\label{tab:classification_report}
\renewcommand{\arraystretch}{1.18}
\setlength{\tabcolsep}{10pt}
\begin{tabular}{lcccc}
\toprule
\textbf{Class} & \textbf{Precision} & \textbf{Recall}
  & \textbf{F1-score} & \textbf{Nb of patients} \\
\midrule
Control & \textbf{1.00} & 0.92 & 0.95 & 6 (46 Images) \\
ALS     & 0.94 & \textbf{1.00} & \textbf{0.97} & 8 (79 Images) \\
\bottomrule
\end{tabular}
\end{table}

Table~\ref{tab:comparison} presents comparison between MRI-based models, HRUS-only baselines, cross-modal learning approaches, and the proposed \textsc{AlignUS} on the same test set, with a column indicating whether each method is deployable using HRUS-only at inference time, which corresponds to the clinically relevant low-resource setting. MRI-based models, including ResNet-18, DenseNet-121 \cite{huang2017densenet}, EfficientNet-B0 \cite{tan2019efficientnet}, and ViT-B/16 \cite{dosovitskiy2021vit}, are marked as not HRUS-deployable ($\times$) since they require MRI inputs at inference, and while they achieve generally strong performance, they serve primarily as upper-bound references. In contrast, HRUS-only baselines are fully deployable ($\checkmark$) but exhibit limited discriminative performance, with balanced accuracy ranging from 0.625 to 0.771 depending on architecture, highlighting the difficulty of learning directly from ultrasound data under limited supervision. Cross-modal methods, including vanilla knowledge distillation \cite{hinton2015distilling} and Fitnets \cite{romero2015fitnets}, also retain HRUS-only inference capability and significantly improve performance (balanced accuracy ~0.839), demonstrating that MRI-guided supervision provides meaningful representational benefits even when MRI is not available at test time. \textsc{AlignUS}, which is also HRUS-deployable ($\checkmark$), achieves the best overall performance among all HRUS-only inference methods, with a balanced accuracy of $0.958$, macro-F1 of $0.963$, and ROC-AUC of $0.990$. This result shows that \textsc{AlignUS} not only surpasses all direct HRUS baselines and existing cross-modal distillation techniques, but also closes a significant portion of the performance gap toward MRI-based models. These findings indicate that MRI-derived features can be transferred into HRUS representations during training, making \textsc{AlignUS} particularly suitable for deployment in low-resource clinical environments where only ultrasound imaging is available at inference time. Notably, \textsc{AlignUS} achieves this performance gain without incurring any additional inference cost relative to a standalone HRUS classifier, since the MRI teacher is discarded after training and the student retains the same ResNet-18 architecture, yielding an inference time of 0.82 ms comparable to the fastest HRUS baseline and substantially lower than ViT-B/16 (4.20 ms), underscoring \textsc{AlignUS}'s suitability for real-time clinical use.

\begin{table}[t]
\centering
\caption{Comparison of MRI-based models, HRUS-only baselines, cross-modal methods,
and \textsc{AlignUS}. The ``Low-resource'' column indicates HRUS-only deployability at inference
time. The inference time of each model is also reported in ms.}
\label{tab:comparison}
\begin{tabular}{l l c c c c c}
\toprule
Modality & Method & Low-resource & Bal. Acc. & F1 & AUC & Time (ms)\\
\midrule
\multirow{4}{*}{MRI}
 & ResNet-18~\cite{he2016}                    & \texttimes & 0.973 & 0.965 & 0.993 & 0.82 \\
 & DenseNet-121~\cite{huang2017densenet}      & \texttimes & 0.905 & 0.898 & 0.934 & 2.35 \\
 & EfficientNet-B0~\cite{tan2019efficientnet} & \texttimes & 0.934 & 0.926 & 0.928 & 1.55 \\
 & ViT-B/16~\cite{dosovitskiy2021vit}         & \texttimes & 0.748 & 0.732 & 0.821 & 4.20 \\
\midrule
\multirow{4}{*}{HRUS}
 & ResNet-18~\cite{he2016}                    & \checkmark & 0.625 & 0.576 & 0.776 & 0.82 \\
 & DenseNet-121~\cite{huang2017densenet}      & \checkmark & 0.771 & 0.757 & 0.865 & 2.35 \\
 & EfficientNet-B0~\cite{tan2019efficientnet} & \checkmark & 0.677 & 0.644 & 0.812 & 1.55 \\
 & ViT-B/16~\cite{dosovitskiy2021vit}         & \checkmark & 0.724 & 0.699 & 0.849 & 4.20 \\
\midrule
\multirow{3}{*}{Cross-modal}
 & Vanilla KD~\cite{hinton2015distilling}     & \checkmark & 0.839 & 0.842 & 0.964 & 0.82 \\
 & Fitnets ~\cite{romero2015fitnets}         & \checkmark & 0.839 & 0.840 & 0.960 & 0.82 \\ 
\rowcolor{gray!10}
 & \textbf{\textsc{AlignUS} (Ours)}                    & \checkmark & \textbf{0.958} & \textbf{0.963} & \textbf{0.990} & \textbf{0.82} \\
\bottomrule
\end{tabular}
\end{table}

\subsection{Embedding Space Analysis}

\begin{figure}[h]
    \centering
    \includegraphics[width=1\linewidth]{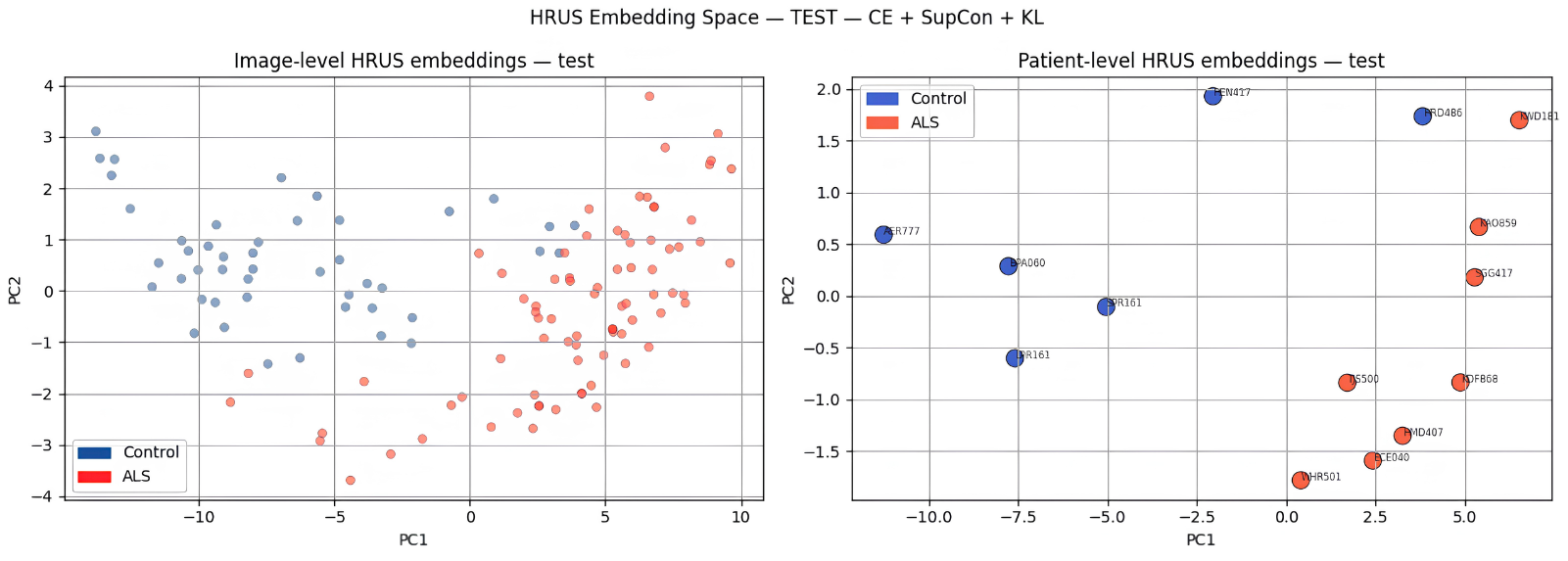}
    \caption{PCA visualization of HRUS embeddings on the held-out test set. Left: image-level embeddings; right: patient-level embeddings. \textcolor{red}{red}: ALS, \textcolor{blue}{blue}: Control.}
    \label{fig:pca_embeddings}
\end{figure}

To further analyze the embedding space learned from HRUS representations, we performed a principal component analysis (PCA) projection~\cite{maaten2008tsne} of the extracted embeddings from the held-out test set, as shown in Figure~\ref{fig:pca_embeddings}. Although some overlap between classes remains at the individual image level, the patient-level aggregated embeddings exhibit a clearer separation between ALS and Control groups, indicating that aggregating multiple HRUS frames enhances representation stability and improves class-level discrimination.


\subsection{Ablation Study}

To evaluate the contribution of each training objective, we progressively removed components from the full \textsc{AlignUS} loss while keeping the preprocessing pipeline, patient-level split, and class-balancing strategy fixed. Results on the held-out test set are reported in Table~\ref{tab:ablation}. We show that performance improves consistently as additional training objectives are introduced.

\begin{table}[t]
\centering
\caption{Patient-level ablation study on the held-out test set.}
\label{tab:ablation}

\footnotesize
\renewcommand{\arraystretch}{1.12}
\setlength{\tabcolsep}{6pt}

\begin{tabular}{lccc}
\toprule
\textbf{Loss function} & \textbf{Bal. Acc.} & \textbf{F1} & \textbf{AUC} \\
\midrule

CE  
& 0.792 & 0.831 & 0.875 \\

CE + KL 
& 0.833 & 0.862 & 0.917 \\

CE + SupCon 
& 0.875 & 0.899 & 0.938 \\

\rowcolor{gray!15}
\textbf{CE + SupCon + KL (\textsc{AlignUS})}
& \textbf{0.958}
& \textbf{0.963}
& \textbf{0.990} \\

\bottomrule
\end{tabular}

\end{table}

\section{Limitations}
\label{sec:limitations}

While \textsc{AlignUS} achieves strong performance for ALS classification from tongue HRUS, the main limitation of this study is the limited size of the available dataset. This constraint arises from the scarcity of publicly available datasets containing both HRUS and MRI acquisitions, which are required for cross-modal knowledge transfer. To mitigate this and reduce the risk of overestimating model performance, we performed four-fold cross-validation using strict patient-level splits, ensuring that images from the same subject were never shared between training and evaluation sets. Nevertheless, the current results represent validation within a single cohort, and further evaluation on larger, multi-center datasets with greater demographic and acquisition variability is required to assess the generalizability and robustness of \textsc{AlignUS} across different clinical environments.

\section{Conclusion}
\label{sec:conclusion}

In this work, we introduced \textsc{AlignUS}, a cross-modal knowledge distillation framework for ALS assessment from tongue HRUS, which leverages MRI-derived anatomical knowledge during training while requiring only ultrasound images at inference time. By transferring structural information from MRI into a portable ultrasound-based classifier, \textsc{AlignUS} enables accurate and accessible disease assessment while preserving the practical advantages of HRUS, including low cost, portability, and suitability for repeated examinations. The proposed framework achieves state-of-the-art performance for ALS classification from tongue ultrasound, demonstrating the potential of cross-modal learning to overcome the limitations of low-resource imaging modalities. Future work will focus on collecting larger and more diverse paired HRUS-MRI datasets and performing external validation across multiple clinical sites to further establish the reliability and clinical applicability of \textsc{AlignUS}.

\bibliographystyle{splncs04}
\bibliography{bib}

\end{document}